\documentclass{article}

\usepackage{PRIMEarxiv}

\usepackage[utf8]{inputenc} 
\usepackage[T1]{fontenc}    
\usepackage{hyperref}       
\usepackage{url}            
\usepackage{booktabs}       
\usepackage{amsfonts}       
\usepackage{nicefrac}       
\usepackage{microtype}      
\usepackage{lipsum}
\usepackage{fancyhdr}       
\usepackage{graphicx}       
\usepackage{orcidlink}
\usepackage{tabularx}
\usepackage{tipa}
\usepackage{subfig}
\usepackage{float}

\title{Phonetically rich corpus construction for a low-resourced language
}

\author{
  Marcellus Amadeus \orcidlink{0009-0002-7777-2562}\\
  Alana AI Research \\
  São Paulo, Brazil\\
  \texttt{marcellus@alana.ai} \\
   \And
  William Alberto Cruz Castañeda
  \orcidlink{0000-0002-9803-1387}\\
  Alana AI Research \\
  São Paulo, Brazil \\
  \texttt{william.cruz@alana.ai} \\
  \And
  Wilmer Lobato
  \orcidlink{0000-0002-3929-1218}\\
  Alana AI Research \\
  São Paulo, Brazil \\
  \texttt{wilmer.lobato@alana.ai} \\
  \And
  Niasche Aquino
  \orcidlink{0000-0003-3006-4163}\\
  Alana AI Research \\
  São Paulo, Brazil \\
  \texttt{niasche.aquino@alana.ai} \\
}

\begin{document}
\maketitle

\begin{abstract}
Speech technologies rely on capturing a speaker's voice variability while obtaining comprehensive language information. Textual prompts and sentence selection methods have been proposed in the literature to comprise such adequate phonetic data, referred to as a phonetically rich \textit{corpus}. However, they are still insufficient for acoustic modeling, especially critical for languages with limited resources. Hence, this paper proposes a novel approach and outlines the methodological aspects required to create a \textit{corpus} with broad phonetic coverage for a low-resourced language, Brazilian Portuguese. Our methodology includes text dataset collection up to a sentence selection algorithm based on triphone distribution. Furthermore, we propose a new phonemic classification according to acoustic-articulatory speech features since the absolute number of distinct triphones, or low-probability triphones, does not guarantee an adequate representation of every possible combination. Using our algorithm, we achieve a 55.8\% higher percentage of distinct triphones -- for samples of similar size -- while the currently available phonetic-rich corpus, CETUC and TTS-Portuguese, 12.6\% and 12.3\% in comparison to a non-phonetically rich dataset.
\end{abstract}


\section{Introduction}
Any acoustic analysis or modeling of a language is based on a speech corpus. This textual dataset might encompass samples of one or more speakers, languages, or varieties, and it is represented on a continuum ranging from spontaneous to stricto sensu laboratory speech \cite{barbosa2015manual}. Considering acoustic feature distribution, statistical learning enables distinct models to recognize or synthesize speech, and identify or compare speakers, allowing the emergence of various applications such as voice-activated personal assistants, automatic speech recognition (ASR) systems, forensics and biometrics usages, and text-to-speech (TTS) synthesis. However, this general structure poses a challenge, especially for low-resource languages. 

Models' performance and features depend on data availability and adequacy regarding speaker and language-specific speech feature representation. For example, during the last few years, speech applications have notably improved their performance, particularly the naturalness and intelligibility of synthesized speech. \cite{Lobato2023} evaluated three promissory TTS models for Brazilian Portuguese (PT-BR) --  considered a low-resourced language. Despite the improvements, results and features available for Chinese and English still outmatch PT-BR and other languages.

With about 230 million native speakers, PT-BR is one of the most widespread languages in the world. However, due to the lack of high-quality speech datasets (with transcription), training deep learning models is a difficult task \cite{Casanova2021b, Neto2015}. To address the issue, techniques are frequently applied to select or develop a corpus encompassing a broad range of phonetic diversity. Such proposals were known as phonetically rich corpus and have been presented for various languages with limited resources, as in \cite{Shulby2018}, \cite{Ahmad2021}, and \cite{Abushariah2012}.

For PT-BR, our low-resource language case study, \cite{Alcaim1992,Seara1994} proposes a list of 200 phonetically balanced sentences. Furthermore, \cite{Cirigliano2005} select 1000 phonetically rich phrases from the CETEN-Folha dataset using genetic algorithms. These works applied a chi-square statistical analysis to compare to a larger sample. More recent research published in \cite{Shulby2018} and \cite{Mendonca2014} analyzed the variability of triphones using \emph{greedy} algorithms. These studies rely on increasing the variety of contextual units (e.g., biphones and triphones) based solely on phonemic transcription, which does not ensure phonetic richness. For example, some triphones are acoustically closer to others ([aua] is more similar to [eua] than [st\textfishhookr]). Thus, it's critical for further phonological analysis to guarantee actual acoustic variability. Another sine qua non-condition, from an applied standpoint,  is a speaker's pronunciation that matches the Grapheme-to-Phoneme converter output.

As described by \cite{Cirigliano2005}, the accuracy of an acoustic model increases with the number of distinct phonetic unit occurrences in a text corpus. Therefore, regardless of numerous studies suggesting phonetically rich and balanced text corpora \cite{Cirigliano2005,Mendonca2014,Shulby2018},  they are still insufficient, and further improvements in speech variability representation have a direct impact on the model's quality.

In this way, the present work proposes a new approach to construct a phonetically rich text corpus for a low-resourced language, Brazilian Portuguese, based on the acoustic-articulatory speech features, and it outlines the methodological aspects of the recording protocol implementation.


\subsection{Initial guidelines}
When constructing or selecting a speech dataset for specific purposes, an aprioristic question emerges: are spontaneous or controlled samples required? The answer will vary according to research and application goals. For instance, training acoustic models for ASR applications requires speech samples similar to its usage context, namely, spontaneous speech samples. In our case, that is to construct a phonetically rich corpus, the lack of control in spontaneous speech does not allow for a careful selection of samples and variables that affect the research goals. Thus, such supervision is vital for the investigation's success.

The laboratory speech -- a strict experimenter control of how phonetic variables are being produced -- has frequently been described as unnatural, overly clear, highly organized, and lacking in rich prosody, notably regarding emotions and communication functions, according to Xu \cite{XU2010329}. However, the author demonstrates that it is more an issue of experimental design rather than the outcome of a fundamental limitation. More crucially, since it allows for systematic control, laboratory speech is essential to understanding the fundamentals of human language – allowing it to produce or optimize new technologies. In contrast, because it is difficult to identify and control the contributing factors to prosodic features, spontaneous speech is less likely to yield valuable findings about language structure.

Based on a reading task, we set a bottom-up approach to ensure complete phonemic coverage of the language -- from phonemes to triphones phonemic classification -- and followed Barbosa's corpora classification proposal \cite{barbosa2012conhecendo}, which includes two variables: experimenter control and textual genre. We control all sentences to be read based on their phonemic coverage and triphones distribution to produce a phonetically rich corpus. For the recording sessions, we set a minimum level of experimenter intervention, resulting in an in-between position on the spectrum of spontaneous to \textit{stricto sensu} laboratory speech. In the case of textual genre, to induce maximum variability within the reading style, we select diverse corpora including different contexts, content, and styles.

From these guiding principles, the following sections outline methodological aspects required to construct a phonetically rich dataset and describe the results obtained by the proposed approach.

\section{Methodology}
\begin{figure*}[!ht]
\begin{center}
    \includegraphics[width=0.9\linewidth]{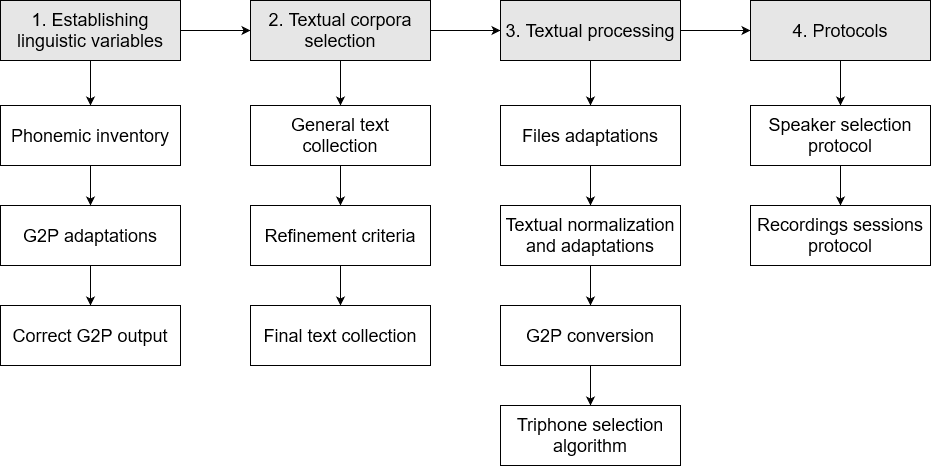}
\end{center}
  \caption{Methodology for building a phonetically rich text corpus for PT-BR.}
\label{metodfig}
\end{figure*}

To build a phonetically rich speech corpus, we must control three broad aspects: language, speaker variables, and recording methods. The proposed methodology, depicted in Figure \ref{metodfig}, for creating a phonetically rich text  for PT-BR is divided into four stages:

\begin{itemize}
    \item Establishing linguistic variables, such as a phonemic inventory for the assumed language variety;

    \item Textual corpora selection, which serves as our source for selecting textual prompts to be recorded in the following stages;

    \item Text processing, ranging from file modifications to sentence selection algorithms;

    \item Speaker selection and recording protocols, which establish criteria for voice selection and discuss recording conditions/tools.

\end{itemize}

These main stages will be discussed in the following sections. 

\subsection{Establishing linguistic variables}
\label{sec:phonetic_inventory} 
When dealing with speech and phonetic features, any approach is subject to the diatopic variables of the language in question. Hence it is critical to assume a specific dialect and a speaker's pronunciation that matches the phonemic transcription.

Along this work, we assumed the São Paulo dialect as our Brazilian Portuguese representative since is the most populous dialectal region \cite{barbosa2004brazilian}. After applying some corrections on the grapheme-phoneme converter \footnote{Phonetic dictionaries for PT-BR can be found in \emph{Aeiuoadô} \cite{Mendonca2014}, \emph{Fala Brasil} group \cite{Batista2022} and Montreal Forced Aligner \cite{Mcauliffe2017}. Moreover, examples of G2P for PT-BR are: \emph{Alinha-PB} \cite{Kruse2021}, Montreal Forced Aligner \cite{Mcauliffe2017} and Phonemizer \cite{Bernard2021}. We adapted -- based on simple phonotactic rules -- available tools to obtain a transcription for our selected dialect.}, the obtained transcription resulted in a phonemic inventory identical to the one described by \cite{barbosa2004brazilian} (Cf. Table \ref{tabtabelaY}).

The consonant inventory abbreviations of Table \ref{tabtabelaY} are described as follows: The place of articulation: BL - Bilabial; LD - Labio-dental; D - Dental; A - Alveolar; PA - Post-alveolar; P - Palatal; V - Velar. The manner of articulation: PL - Plosive; Afr - Affricate; Na - Nasal; Tap - Tap; Fri - Fricative; Lat - Lateral.

\begin{table}[!ht]
\begin{center}
\begin{tabularx}{\columnwidth}{|l|X|X|X|X|X|X|X|}
    \hline
    & \textbf{BL} & \textbf{LD} & \textbf{D} & \textbf{A} & \textbf{PA} & \textbf{P} & \textbf{V}\\
    \hline\hline
    \textbf{Pl} & p b &  & t d & & & & k g\\
    \hline
    \textbf{Afr} & & & \textteshlig \textdyoghlig & & & &\\
    \hline
    \textbf{Na} & m & & n & & & \textltailn &\\ 
    \hline
    \textbf{Tap} & & & & \textfishhookr & & & \\
    \hline
    \textbf{Fri} & & f v & & s z & \textesh\textyogh & & \textgamma\\
    \hline
    \textbf{Lat} & & & & l & & \textturny & \\
    \hline
\end{tabularx}
\caption{Consonant inventory of São Paulo dialect for Brazilian Portuguese}
\label{tabtabelaY}
 \end{center}
\end{table}

On the other hand, the vowel inventory abbreviations of Table \ref{tab:tabelaZ} are described as follows:
In terms of tongue advancement: Frt - Front; NFrt - Near Front; Ce - Central; NBck - Near Back; Bck - Back. 
The vowel height: Cls - Close; NCls - Near Close; MCls - Close-mid; MOpn - Open-mid; NOpn - Near-Open; Opn - Open.

\begin{table}[!ht]
\begin{center}
\begin{tabularx}{\columnwidth}{|l|X|X|X|X|X|}
    \hline
    & \textbf{Frt} & \textbf{NFrt} & \textbf{Ce} & \textbf{NBck} & \textbf{Bck}\\
    \hline\hline
    \textbf{Cls} & i & & & & u\\
    \hline
    \textbf{NCls} & & I & & \textupsilon &\\
    \hline
    \textbf{MCls} & e & & & & o\\
    \hline
    \textbf{MOpn} & \textepsilon & & & & \textopeno\\
    \hline
    \textbf{NOpn} & & & \textturna & &\\
    \hline
    \textbf{Opn} & a & & & &\\
    \hline
\end{tabularx}
\caption{Vowel inventory of São Paulo dialect for Brazilian Portuguese}
\label{tab:tabelaZ}
 \end{center}
\end{table}

\subsection{Textual corpora selection}
\label{sec:data-collection}
In this section, we explained our data collection method, justifying everything from inputting a textual corpus to generating phonetically rich sentences. We carried out a comprehensive collection of textual corpora available for PT-BR, encompassing multiple contexts to address the issue of managing its probable influence on speech variability, particularly prosody. Since we are dealing with a low-resourced language, we have established a lenient inclusion criterion to allow for the inclusion of a large number of datasets, that is being a PT-BR-specific text. Our initial collection resulted in 43 corpora included.

\subsubsection{Exclusion criteria}
\label{sec:inclusion-exclusion}
To establish our final corpus selection, we defined the following exclusion criteria, adjusted by an exploratory analysis of each dataset. We have set the following desiderata:

\begin{itemize}
\item Inappropriate content, such as slang, profanity, and sexual content;  \item Minimum text size: at least 5000 tokens per context;  
\item High percentage of misspellings, informal abbreviations, and other non-standard writing characteristics;  
\item Text date: from 1990 to 2023. A broad time interval was selected, as PT-BR is a language with few resources; 
\item Text annotated with only one genre and/or domain.
\end{itemize}

There were four types of communicative contexts and goals identified: news from Brazil (BN), chatbot (CH), spontaneous speech transcripts (SST), and online media (OM). According to the literature, existing approaches choose phrases based on only one context, with CETEN-Folha as the primary text corpus employed. We select the following text corpora using exclusion criteria to assure data quality and context variability: BrWac \cite{wagner2018brwac}, CETEN-Folha \cite{ceten2014}, Globo \cite{Leite2020}, CORAA \cite{junior2021coraa}, Alana, and Chatterbot corpus. Table \ref{tokens} describes word tokens and estimated durations (as described in section \ref{sec:corpus_selection})  for each selected corpus.

\begin{table}[!ht]
\begin{center}
\begin{tabularx}{\columnwidth}{|l|X|X|}
    \hline
    \textbf{Corpus text} & \textbf{Tokens} & \textbf{Duration}\\
    \hline\hline
    brWac & $2,680'000.000$ & $496.300 $h \\
    \hline
    CETEN-Folha & $33'000.000 $ & $6.100 $h \\
    \hline
    Globo & $180.000$ & $20 $h \\
    \hline
    CORAA & $56.412 $ & $290,77 $h \\
    \hline
    Internal dataset & $12.515$ & $2,3 $h \\
    \hline
    Chatterbot & $5.709$ & $1 $h \\
    \hline
    \textbf{Total} & \textbf{2.713.000.000} & \textbf{502715 h} \\
    \hline
\end{tabularx}
\caption{Number of tokens per corpus text and estimated recordings durations.}
\label{tokens}
 \end{center}
\end{table}

\subsection{Textual processing}
\label{sec:text_processing}

After the collection of corpora (as described in Section \ref{sec:data-collection}), each file is adapted to a standard format, and sentences are extracted and classified into declarative, interrogative, and exclamatory sentences. First, we set a corpus size of an estimated 20 hours to select a matching number of sentences. Then, these phrases are then phonetically transcribed and processed until they reach a final stage of triphone distribution-based selection and classification. Each processing stage is described in this section.

\subsubsection{Corpora size selection}
\label{sec:corpus_selection}
There are two primary parameters to consider when determining the number of sentences needed to obtain a specific speech sample size: average speech rate and sentence length.

According to the literature, Brazilian Portuguese has an average speech rate of 6 syllables per second, which has been tested for different varieties \cite{gonçalves_2017,meireles2009papel}. As stated by \cite{Mendonca2014}, long sentences, despite their high richness value, can be too complex for a recording script and might induce speech disfluencies (i.e. false starts and self-correction). On the other hand, short sentences selected by their algorithm were usually only nominal (titles and topics) and were not suited for sentence prompts aiming for a natural representation of speech. This issue led us to establish minimum and maximum triphones per sentence.

Within this scope, suppose sentences with 10 to 20 words with an average of 4 syllables, as observed by \cite{viaro2007analise}. By assuming a speech rate of 6 syl/s, for 20 hours of recording for each speaker, our set must contain at least 10000 sentences. As a precautionary measure, we choose an additional set of 2000 sentences to be used if the duration is less than a predetermined limit.

The selected set is composed of 10000 phonetically rich sentences divided into 6000 declarative, 3000 interrogative, and 1000 exclamative \cite{Casanova2022}. This proportion is followed within a variety of contexts, including newspapers and online shopping, to account for higher variability in speaking styles and genres, resulting in a more holistic prosodic feature distribution.

\subsubsection{Sentences selection}
\label{sec:sentences_selection}

After grapheme-phoneme conversion, we start the selection stage. The first step is to ensure complete phonemic coverage and a minimum threshold for each phoneme of the assumed variety. For this, we define an initial selection algorithm that accepts a new phrase only if it contributes to a new occurrence of a phoneme with a frequency below the established threshold. Then, we proceed with the classification of the assumed contextual unit, that is, the triphones.

Triphones are the most typical minimal unit used to create a phonetically rich corpus for PT-BR \cite{Cirigliano2005,nicodem}. The variety of sets is usually represented by the total number of triphones. We took a step further, assuming a classification with acoustic-articulatory motivation and presenting the distribution of triphones categorized into vocoids/contoids. "Vocoids are sounds produced with a free central passage in the oral cavity, all other sounds are contoids" \cite{aubert1976phonetic}.

The fundamental prosodic characteristics of speech, especially rhythm, are oscillatory and suitably modeled by a dynamic system \cite{barbosa2006incursoes} -- mathematically, it describes the relationship of a point in geometric space as a function of time. Jaw movement has been found to organize consonant gestures in this way around the \cite{rhardisse1995mandible} vowels. This is the basis of our decision to represent the distribution of triphones as vocoid/contoid units. As per the example below:

\begin{itemize}
\item VVV - Pesquis$\vert$a é u$\vert$ma coisa que muda toda hora. - /\textturna \textepsilon u/

\item CCC - Isso é muito e$\vert$xtr$\vert$emo. - /st\textfishhookr/

\item CVV - Bibliotecas foram inauguradas em territó$\vert$rio$\vert$ americano. - /\textfishhookr \textsci \textupsilon/

\item VCV - $\vert$Atu$\vert$almente, esse é o limite. - /atu/

\item VVC - Em Flor$\vert$ian$\vert$ópolis, fez dois graus celsius no domingo. - /i\textturna n/ 

\item VCC - Ele está exultante desde que viro$\vert$u pr$\vert$esidente. -/\textupsilon p\textfishhookr/

\item CVC - $\vert$No t$\vert$otal, serão chamados vinte e seis mil candidatos. - /n\textupsilon t/ 

\item CCV - A mãe de todas a$\vert$s re$\vert$formas é a reforma política. - /s\textgamma e/ 
\end{itemize}

The classification proposed in our work is fundamental because the absolute number of distinct triphones or even the selection of low-probability triphones does not capture/guarantee an adequate representation of all possible combinations. This has a significant impact on less likely sequences such as VVV and CCC, resulting in a corpus that is not as phonetically rich as intended.

The selection algorithm is applied in batches and it is based on a set of thresholds that capture such fundamental characteristics of speech structure and, thus, allow forming a locally optimal set in terms of phonetic richness. This operates according to the pseudocode below:

\begin{footnotesize}
\begin{verbatim}
1. Textual input:
     - Text cleaning;
     - Tokenization of the sentence;
     - Sentence type classification;
2. Grapheme-phoneme conversion;
3. Classification of triphones;
4. Calculation of the contribution of 
new triphones;
5. Calculation of linguistic thresholds;
6. If higher than established thresholds:
         Add sentence;
         Recalculate triphones inventory;
7. Add batch results;
8. Return to step 3;
\end{verbatim}
\end{footnotesize}

After applying the algorithm, an additional adaptation step was performed with regular expressions and \textit{part-of-speech tagging} algorithms. This phase aimed to avoid triphone combinations that are not allowed by the phonotactic rules of Brazilian Portuguese. Above all, dealing with foreign words and proper names that deviate from spelling rules.

\subsection{Protocols: speakers and recordings}
Finally, we established a speaker selection protocol (age, dialect, etc.) and recording conditions (format, quantization rate, etc.). Based on a set of criteria, we aim to ensure, above all, that the voice is suitable for TTS applications, that the pronunciation corresponds to that generated by the grapheme-phoneme converter (since all phonetic richness depends on this compatibility), and that the samples will be collected properly to result in good quality audios also in the synth.

The main range of interest in the speech frequency domain is within 50 Hz (for instance, the fundamental frequency of an adult male) up to 15 kHz (e.g. first formant of alveolar fricatives of small children) \cite{barbosa2015manual}.  By the Nyquist theorem\footnote{“The Nyquist theorem specifies that a sinuisoidal function in time or distance can be regenerated with no loss of information as long as it is sampled at a frequency greater than or equal to twice per cycle” \cite{lindon2016encyclopedia}.} to have adequate information, we need a sampling rate of at least 30 kHz. For samples with high-frequency resolution, we propose a value of 48 kHZ (DAT standard) or 44.1 kHZ (CD standard) with at least a 32-bit resolution \cite{Casanova2022}. Audio files must be in WAV (Waveform Audio File) since its uncompressed format retains the highest audio quality possible.

The frequency response and type are the two most important factors to consider when selecting the microphone. We need a uniform (flat) frequency response within the human audible range of 20 Hz to 20 kHz for a natural-sounding audio sample without acoustic modifications. To avoid capturing environmental noise and having a speech sample with high detail richness and accuracy, we recommend using a unidirectional (cardioid) condenser microphone with a smaller diaphragm.

Hence, we need to maximize the robustness of acoustic signals by achieving a high signal-to-noise ratio to guarantee recordings and synthesized speech quality. In addition to microphone and recording room settings, we require a high gain recording level with no clipping/overloading of the signal. Since our goal is to capture speech samples as naturally as possible, no automatic gain control or recording effects/modifications should be used. 

Another influencing factor is intensity, which varies according to the Law of Inverse Squares and decreases at a rate proportional to the square of the distance from the sound source. To preserve high gain, the speaker should maintain a distance of 20 centimeters from the microphone, and specific adjustments should be performed. In summary, the following conditions are required to record high-quality and natural audio samples:

\begin{itemize}
    \item File format: Waveform Audio File (WAV);

    \item Sampling rate:  48 kHz (DAT standard) or 44.1 kHz (CD standard);

    \item Bits resolution: at least 32-bits;

    \item Unidirectional (cardioid) condenser microphone with a smaller diaphragm;

    \item Uniform (flat) frequency response within the human audible range of 20 Hz to 20 kHz 

    \item Speaker must keep a distance of 20 cm from the microphone at all times;

    \item No presence of reverberation and/or background noise;

    \item Acoustically treated recording room; 

    \item High gain recording level with no clipping/overloading of the signal;

    \item No automatic gain control (ACG) or recording effects/modifications.
\end{itemize}

\section{Discussion and results}
\label{sec:discussion}
Many speech applications require the higher variability of phonetic contexts possible for each of its minimal units (phonemes, syllables, etc) so that the statistical learning method can create accurate distributions for the model states \cite{SIG-001}. Capturing speech variability is even more crucial -- for training different models with minimum adequate data -- when there is a lack of data, as for low-resourced languages. 

\begin{figure}[!ht]
\begin{center}
    \includegraphics[width=0.65\linewidth]{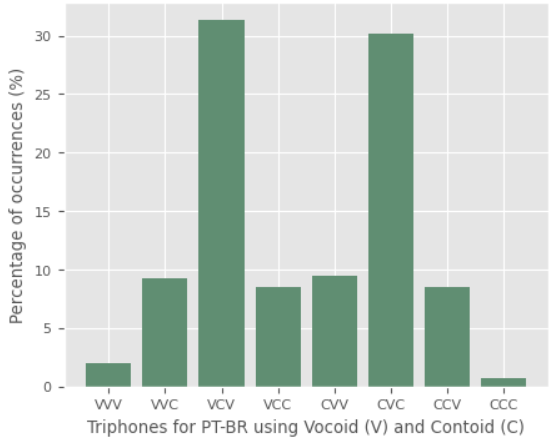}
\caption{Phonemic histogram of the CETUC corpus text by considering a vocoid-contoid triphone classification.}
\label{fig:apon2}
\end{center}
\end{figure}

\begin{figure}[!ht]
\begin{center}
    \includegraphics[width=0.65\linewidth]{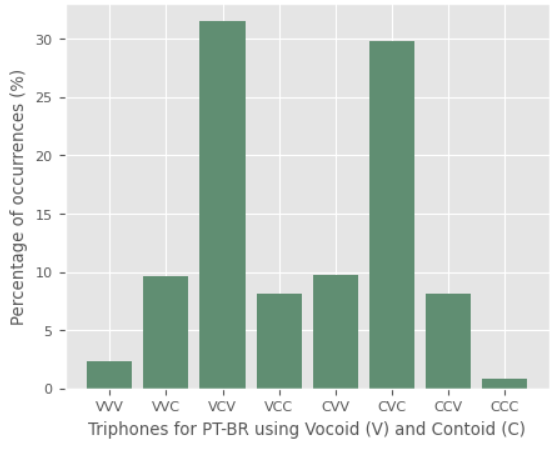}
\caption{Phonemic histogram of the CETUC corpus text by considering a vocoid-contoid triphone classification.}
\label{fig:apon3}
\end{center}
\end{figure}

By applying our selection algorithm, we observed that the balancing of categories is intrinsic to the structure of the language (\textit{Cf.} Figures \ref{fig:apon2} and \ref{fig:apon3}). Phonotactic constraints apply to contoid-vocoid triphone distribution on a larger corpus. In this way, the phonetic richness resides in increasing the ratio of distinct triphones in relation to the total value -- a logic that applies to the proposed categories.

It is evident that the number of new triphones per sentence tends to be 0. However, it is fundamental to define the number of sentences necessary for this to occur since it is not very productive to evaluate more sentences after such saturation. Allowing, therefore, to determine the ideal size of the batches to be used.

\begin{figure}[!ht]
\begin{center}
    \includegraphics[scale=0.48]{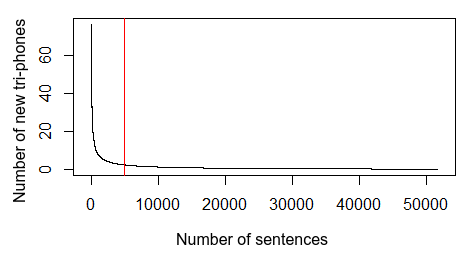} 
    \caption{Changepoint analysis for the variance of new triphones per sentence.}
    \label{change}
\end{center}
\end{figure}

For this, from the cumulative average of new triphones per sentence, we calculated the variance stabilization. Given its skewed distribution, we used the nonparametric cumulative sum of squares test to detect significant changes in variance \cite{inclan1994use}. We verified that the number of new triphones stabilizes around 5000 sentences (\textit{Cf.} Figure \ref{change}). 

\begin{figure}[!ht]
\begin{center}
    \includegraphics[width=\linewidth]{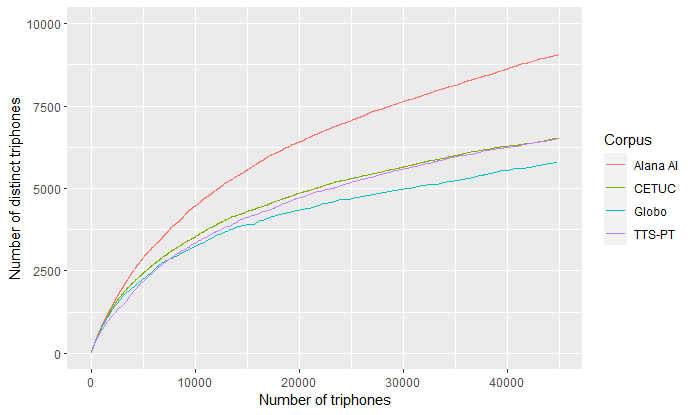} 
    \caption{Corpora new triphones per sentence.}
    \label{ntric}
\end{center}
\end{figure}

Considering current SOTA datasets available for PT-BR, a low-resourced language, our approach resulted in a higher phonetic richness within samples of the same size, as shown in Figure \ref{ntric}. In this way, the Alana AI corpus conveys more information about language structure, and it is more suited to acoustic modeling tasks.

In addition to presenting a greater variability of contexts, encompassing declarative, interrogative, and exclamatory sentences, the results for samples with the same length as CETUC (about 45 thousand triphones), which is also a phonetically rich corpus, obtained much higher values of distinct triphones per total. In a comparison with Globo, a non-phonetically rich dataset, TTS Portuguese and CETUC have, respectively, 12.6\% and 12.3\% more distinct triphones. Alana AI has 55,8\%.

\begin{table}[!ht]
\begin{center}
\begin{tabularx}{\columnwidth}{|l|X|X|X|}
    \hline
    \textbf{Corpus} & \textbf{Distinct Tri.} & \textbf{Total Tri.} & \textbf{Ratio}\\
    \hline\hline
    Alana AI & 9088 & 45360 & $20.04\%$ \\
    \hline
    CETUC & 6551 & 45602 & $14.36\%$ \\
    \hline
    TTS-Portuguese & 6567 & 45557 & $14.41\%$ \\
    \hline
    Globo &  5832 & 45504 & $12.81\%$ \\
    \hline
\end{tabularx}
\caption{Distinct triphones per samples.}
 \end{center}
\end{table}

\begin{table}[!ht]
\begin{center}
\begin{tabularx}{\columnwidth}{|l|X|X|X|}
    \hline
    \textbf{Corpus} & \textbf{Distinct Tri.} & \textbf{Total Tri.} & \textbf{Ratio}\\
    \hline\hline
    Alana AI & 15359 & 348824 & $4.40\%$ \\
    \hline
    TTS-Portuguese & 11108 & 326900 & $3.39\%$ \\
    \hline
    Globo &  12801 & 834714 & $1.53\%$ \\
    \hline
\end{tabularx}
\caption{Distinct triphones per corpora.}
 \end{center}
\end{table}

The superiority of the Alana AI corpus is not restricted to the percentage of distinct triphones. As noted earlier, it is essential to observe the triphonic distribution by category. Thus, it is noted that there is greater variability for all proposed classifications.

\begin{table}[!ht]
\begin{center}
\begin{tabularx}{\linewidth}{|X|c|c|c|c|}
    \hline
    \textbf{Type} & \textbf{Alana AI} & \textbf{CETUC} & \textbf{TTS-PT}  & \textbf{Globo}  \\
    \hline\hline
    VVV & $38,0\%$&  $27,8\%$ & $7,7\%$ & $4\%$ \\
    \hline
    VVC & $28,3\%$ & $21\%$ & $5,3\%$ & $2,3\%$ \\
    \hline
    VCV & $15,5\%$ &$12,2\%$ & $2,5\%$& $1,1\%$ \\
    \hline
    VCC & $19,7\%$ &  $11,6\%$ & $3,1\%$ & $1,6\%$ \\
    \hline
    CVV & $22,8\%$ &  $16,4\%$ & $4,3\%$ & $1,9\%$ \\
    \hline
    CVC & $19,3\%$ & $14,3\%$ & $3,1\%$ & $1,3\%$ \\
    \hline
    CCV &$20,3\%$& $11,9\%$& $3,2\%$ & $1,6\%$ \\
    \hline
    CCC & $39,5\%$ & $22,9\%$ & $8,2\%$& $3,9\%$ \\
    \hline
\end{tabularx}
\caption{Comparison of the percentage of distinct triphones per category between corpora.}
 \end{center}
\end{table}

\section{Conclusion}
\label{sec:conclusion}

This work aimed to construct a phonetically rich corpus text for a low-resourced language, that is Brazilian Portuguese, which has significantly fewer language resources compared to English and Chinese \cite{Casanova2021b} \cite{Neto2015}. Throughout this study, we proposed a methodology from the collection of data, the grapheme-to-phoneme converter, triphone computing, and the sentence selection algorithm, aiming to only consider the sentences that add new triphones into the CV categories. Results demonstrated that the corpus text obtained is richer in triphones than CETUC, TTS-Portuguese \cite{Casanova2021b} and Globo \cite{Leite2020}, which were generated in recent years to train TTS models for Brazilian Portuguese.

%


%
%
%
%
%

\section{Acknowledgements}
This research is supported by the National Council for Scientific and Technological Development (CNPq/MCTI/SEMPI Number 021/2021 RHAE - 351436/2022-7), Ministry of Science, Technology and Innovation, and Alana AI.



\section{Bibliographical References}\label{reference}

\bibliographystyle{unsrt}
\bibliography{references}

\end{document}